\documentclass[conference]{IEEEtran}

\usepackage{cite}
\usepackage{amsmath}
\usepackage{algorithm}
\usepackage{algpseudocode}
\usepackage{amsmath}
\usepackage{graphicx}
\usepackage{textcomp}
\usepackage{xcolor}
\usepackage{hyperref}

\begin{document}
\title{MoonMetaSync: Lunar Image Registration Analysis}

\author{\IEEEauthorblockN{Ashutosh Kumar}
\IEEEauthorblockA{\textit{School of Information} \\
\textit{Rochester Institute of Technology}\\
Rochester, NY, USA \\
ak1825@rit.edu}
\and
\IEEEauthorblockN{Sarthak Kaushal}
\IEEEauthorblockA{\textit{School of Information} \\
\textit{Rochester Institute of Technology}\\
Rochester, NY, USA \\
sk4858@rit.edu}
\and
\IEEEauthorblockN{Shiv Vignesh Murthy}
\IEEEauthorblockA{\textit{School of Information} \\
\textit{Rochester Institute of Technology}\\
Rochester, NY, USA \\
sm2678@rit.edu}}

\maketitle

\begin{abstract}
This paper compares scale-invariant (SIFT) and scale-variant (ORB) feature detection methods, alongside our novel feature detector, IntFeat, specifically applied to lunar imagery. We evaluate these methods using low (128x128) and high-resolution (1024x1024) lunar image patches, providing insights into their performance across scales in challenging extraterrestrial environments. IntFeat combines high-level features from SIFT and low-level features from ORB into a single vector space for robust lunar image registration. We introduce SyncVision, a Python package that compares lunar images using various registration methods, including SIFT, ORB, and IntFeat. Our analysis includes upscaling low-resolution lunar images using bi-linear and bi-cubic interpolation, offering a unique perspective on registration effectiveness across scales and feature detectors in lunar landscapes. This research contributes to computer vision and planetary science by comparing feature detection methods for lunar imagery and introducing a versatile tool for lunar image registration and evaluation, with implications for multi-resolution image analysis in space exploration applications\footnote{The code and dataset for IntFeat and SyncVision are available here: \href{https://github.com/ashu1069/MoonMetaSync}{https://github.com/ashu1069/MoonMetaSync}}.
\end{abstract}

\section{Introduction}
Image registration is aligning two or more images of the same scene, captured at different times, from varying viewpoints, or by different sensors such as high- and low-resolution payloads. This technique is essential for ensuring consistency and accuracy when analyzing and comparing multiple images, making it foundational in fields like remote sensing, medical imaging, and computer vision.\\[5pt]
Feature detection is a crucial step in image registration, with algorithms like SIFT (Scale-Invariant Feature Transform) and ORB (Oriented FAST and Rotated BRIEF) being widely used. SIFT has long been favored for its robustness in detecting distinctive key points and its ability to handle scaling, rotation, and illumination changes. However, SIFT's computational complexity can be a bottleneck, particularly in real-time applications.\\[5pt]
ORB improves upon SIFT by offering a faster and more efficient alternative. It combines the FAST keypoint detector with the BRIEF descriptor, adding rotational invariance and robustness to noise. ORB also incorporates orientation information, making it rotation-invariant like SIFT, while its binary BRIEF descriptor ensures a significantly lighter computational load. This makes ORB suitable for real-time applications such as feature matching in robotics and mobile devices, where speed and efficiency are critical. However, while ORB is computationally faster, it may sacrifice some of the rich feature representation that SIFT provides.\\[5pt]
Our contribution, \textbf{IntFeat}, aims to bridge this gap by taking the best elements from both SIFT and ORB. IntFeat retains the richness of SIFT’s features while optimizing for the speed and efficiency of ORB, making it suitable for real-time applications without compromising on the quality of feature extraction.
%%%%%%%%%%%%%%%%%%%%%%%%%%%%%%%%%%%%%%%%%%%%%%%%%%%%%%%%%%
\section{Related Work}
Image calibration of satellite imagery is crucial for ensuring data quality and enabling accurate scientific analysis across various applications in remote sensing and earth observation. Broadly image calibrations of two types radiometric\cite{b3} and geometric, radiometric corrects for variation in pixel sensitivity and converts raw digital numbers into meaningful physical units. The latter is used to correct for lens distortions and determine camera parameters. Further, geometric calibrations can be of two types intrinsic (determines internal camera parameters like focal length and optical center) and extrinsic (determines camera position and orientation in 3D space). Geometric calibration is used in different tasks like digital breast tomosynthesis (DBT) \cite{b4}, systems with inconsistent imaging capability\cite{b5}, Camera-projector 3D imaging systems\cite{b6}, etc. Geometric calibrations apply different approaches: projection matrix approach\cite{b4}, uncertainty estimation\cite{b5}, simultaneous multi-sensor calibration\cite{b7} (calibrating multiple sensors and determining their misalignment simultaneously), line-based iterative calibration\cite{b8} and simulations camera-projector calibration \cite{b6}. \\[5pt]
Geometric calibration precedes image registration in many applications and provides the necessary parameters for accurate image registration, especially in multi-camera or 3D reconstruction scenarios. Image registration geometrically aligns two or more images of the same scene. This alignment is crucial for various applications in computer vision, medical imaging, remote sensing, and more. The process typically involves detecting and matching distinctive features across images and then estimating the transformation needed to align them. Early approaches included rigid transformations (rotation, scaling, and translation)\cite{b9} and manual registration. Further, these were improved upon by Voxel intensity methods\cite{b10}\cite{b12}, Multi-modality registration\cite{b11}, and methods involving non-linear relationships. But the breakthrough came in the mid-1990s when researchers started using methods involving joint histograms and mutual information\cite{b13}. Modern approaches that used atlas-based registration\cite{b14}, machine learning, and multi-modal\cite{b14} fusion.\\[5pt]
A critical step in image registration is the detection and description of key features in the images. Two popular algorithms for this purpose are SIFT\cite{b2} and ORB\cite{b16}. SIFT extracts distinctive key points and generates robust descriptors that are invariant to image scaling, and rotation, and partially invariant to changes in illumination and viewpoint. The method consists of four major steps: scale-space extrema detection, keypoint localization, orientation assignment, and descriptor generation. SIFT is widely used in tasks such as object recognition, image stitching, and 3D reconstruction due to its accuracy and resilience to transformations. Though computationally intensive, SIFT remains a popular choice for feature-based image registration.
\section{Methodology}
\begin{figure*}
    \centering
    \includegraphics[width=0.9\linewidth]{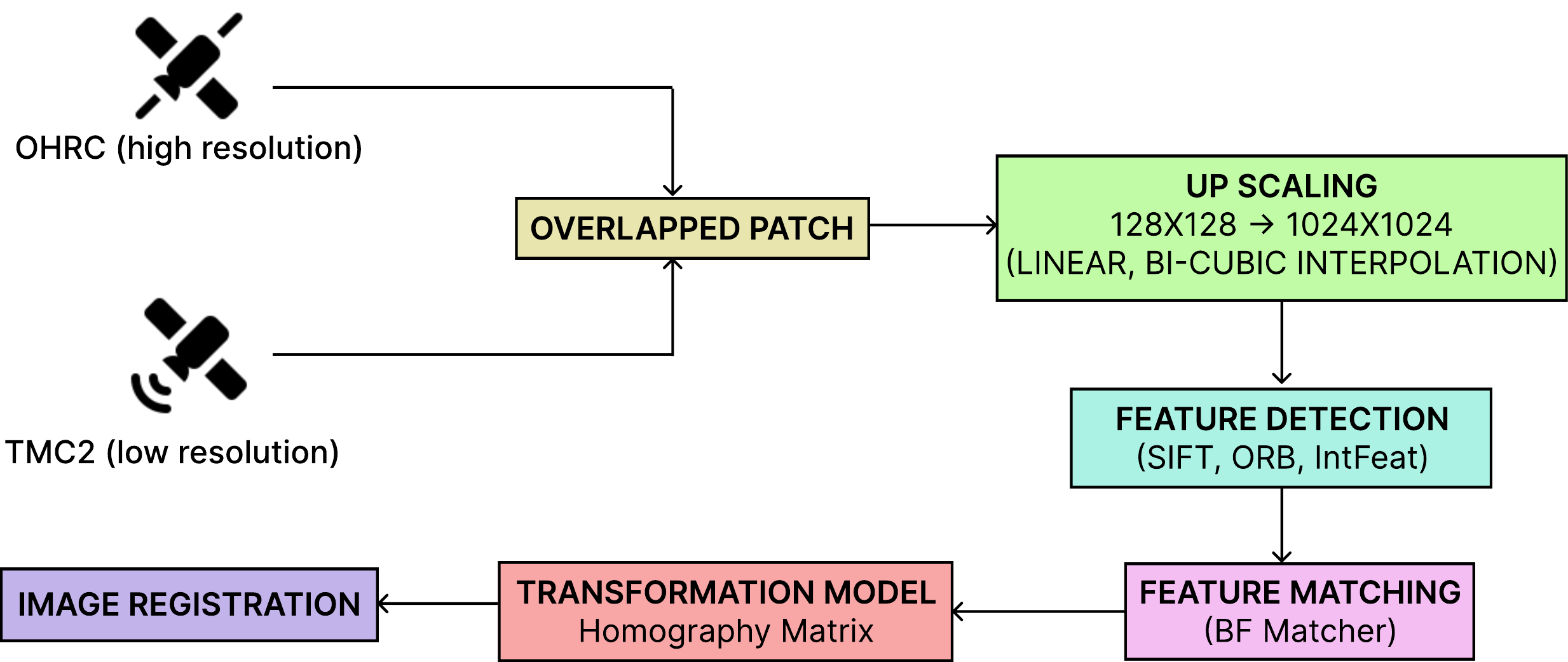}
    \caption{Overall methodology of image registration in the SyncVision framework, showcasing the sequential process of aligning TMC-2 and OHRC patches. This includes feature extraction using advanced algorithms, keypoint matching, and interpolation techniques (both bi-linear and bi-cubic) to achieve optimal alignment. The flow demonstrates how these components work together to enhance image quality and accuracy in lunar terrain analysis, ensuring precise registration of diverse image data.}
    \label{fig:framework}
\end{figure*}
\subsection{Dataset}
The dataset was acquired from ISRO PRADAN servers and includes imagery from the Chandrayaan-II orbiter’s two imaging payloads: Terrain Mapping Camera (TMC-2) and Orbiter High-Resolution Camera (OHRC). TMC-2 provides a resolution of 5 meters, while OHRC offers a much finer resolution of approximately 30 cm. Although TMC-2 has covered more than half of the lunar surface, its images are of lower resolution compared to the limited but higher-resolution coverage provided by OHRC.
\begin{figure}
    \centering
    \fbox{\includegraphics{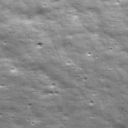}}
    \caption{TMC-2 patch (Size: 128x128)}
    \label{fig:tmc2}
\end{figure}
\begin{figure}
    \centering
    \fbox{\includegraphics[width=0.8\linewidth]{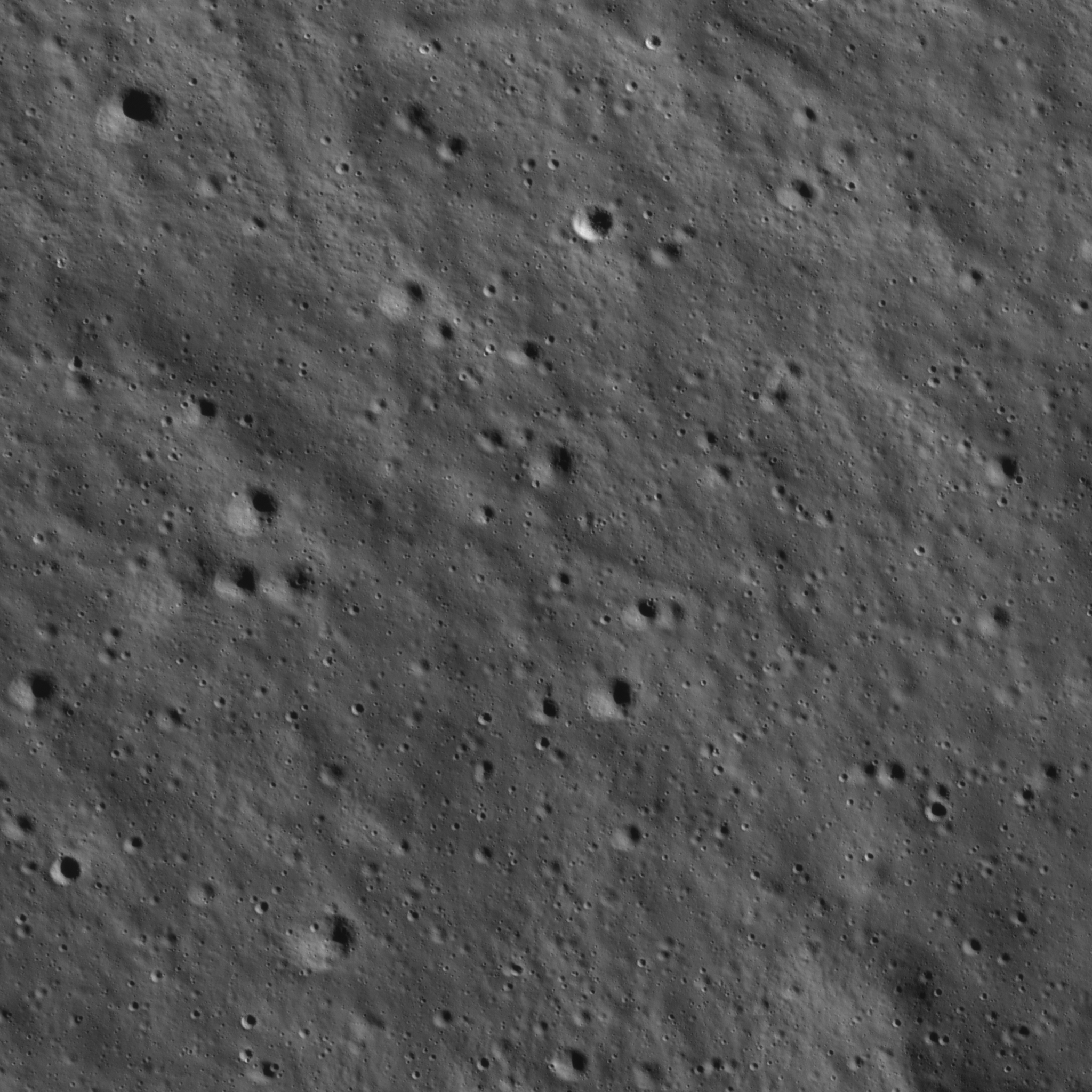}}
    \caption{OHRC patch (Size: 1024x1024)}
    \label{fig:ohrc}
\end{figure}
\begin{figure}
    \centering
    \includegraphics[width=0.6\linewidth]{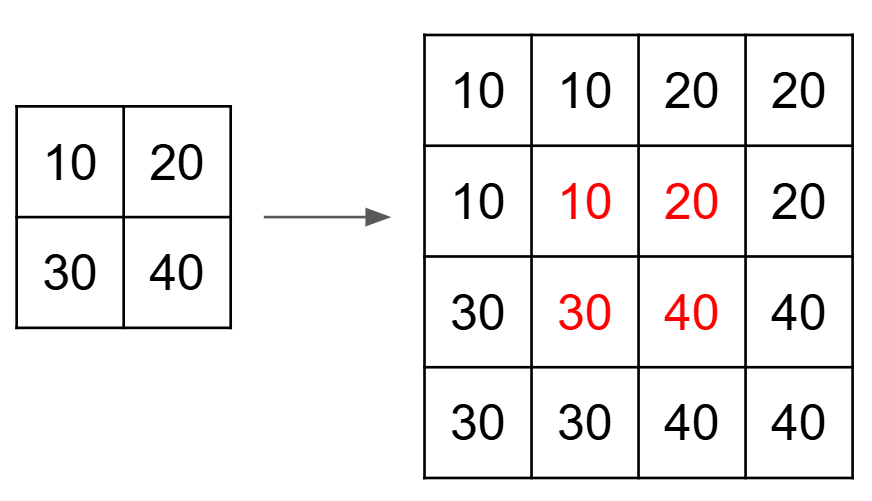}
    \caption{Bi-linear interpolation, calculates the value of a new pixel based on a weighted average of the four nearest pixels in the original image.}
    \label{fig:bilinear}
\end{figure}
\begin{figure}
    \centering
    \includegraphics[width=0.8\linewidth]{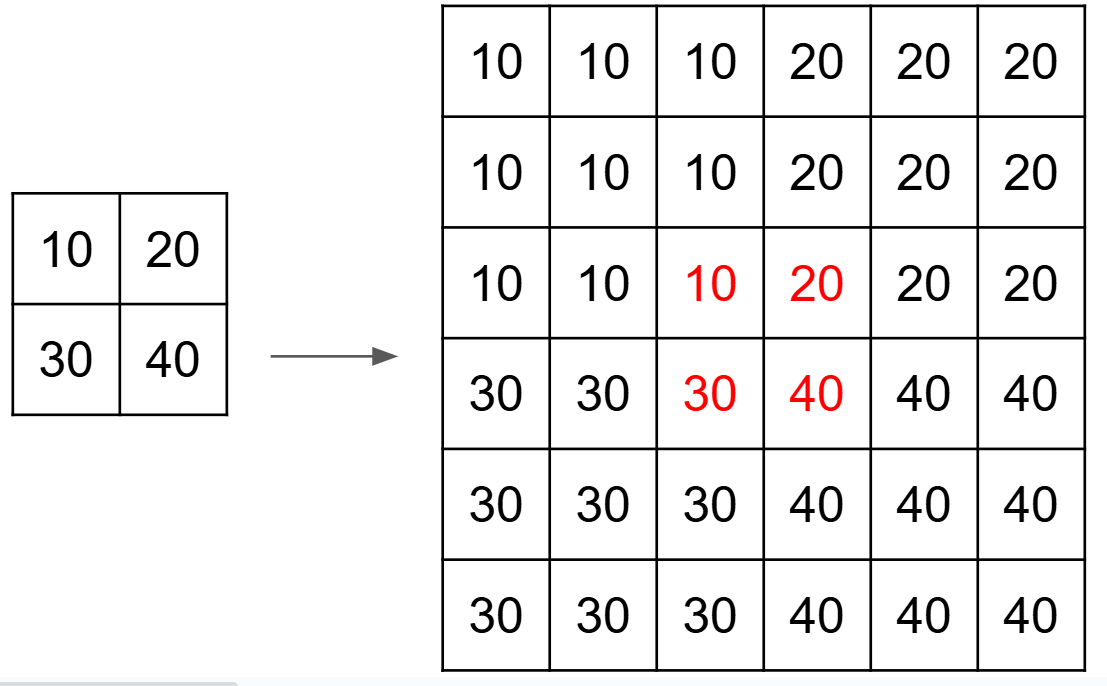}
    \caption{Bi-cubic interpolation, calculates the value of a new pixel based on a weighted average of the 16 nearest pixels in the original image}
    \label{fig:bicubic}
\end{figure}
\subsubsection{Matching OHRC and TMC-2 Images}
To leverage the strengths of both imaging payloads, we aligned the OHRC images with the corresponding regions in the TMC-2 dataset. Using the \texttt{pygeodesy} library, we matched the corner coordinates of OHRC images within the broader TMC-2 image frames. Through this process, 6 unique TMC-2 images and 18 OHRC images were generated, ensuring that the two datasets aligned geographically.
\subsubsection{Image Preprocessing}
The preprocessing steps began with interpolating pixel coordinates around the corner points of OHRC within the corresponding TMC-2 images. We defined a bounding box for the common patch between both image sets, masking the non-overlapping areas.
In total, 16 OHRC images were matched with 6 TMC-2 images. We used geodesic interpolation to map pixel coordinates at every 100th pixel inside the bounding box for both image types. The Haversine distance was calculated between the corner coordinates of OHRC and each pixel inside the TMC-2 bounding box to ensure precise spatial matching. Finally, affine transformations were applied to the OHRC images to align them accurately with the TMC-2 images, ensuring that both datasets correctly represented the same lunar terrain patches.

\subsection{Image Registration}
After collecting and processing the dataset from the OHRC and TMC-2 satellite payloads, we concentrated on upscaling methods, specifically bilinear (Figure \ref{fig:bilinear}) and bicubic (Figure \ref{fig:bicubic}) interpolation techniques, which played a critical role in our subsequent analysis. Bilinear interpolation uses the 4 nearest neighbors to estimate pixel values, while bicubic interpolation utilizes a 4x4 neighborhood for smoother and more accurate pixel value estimation.

\subsubsection{Feature Detection}
In our analysis, we explored various feature detection methods and developed a novel approach, \textbf{IntFeat}, which combines the strengths of existing techniques such as SIFT and ORB.\\[5pt] 
% \textbf{SIFT (Scale-Invariant Feature Transform)}
% \begin{itemize}
%     \item \textit{Keypoint Detection}: SIFT uses the Difference of Gaussians (DoG) to detect scale-space extrema.
%     \item \textit{Keypoint Localization}: A contrast threshold of 0.03 was applied, discarding low-contrast and edge keypoints to retain robust features.
%     \item \textit{Keypoint Descriptor}: A 16x16 neighborhood around each keypoint was used to generate descriptors that capture local image patterns.
% \end{itemize}
% \textbf{ORB (Oriented FAST and Rotated BRIEF)}
% \begin{itemize}
%     \item \textit{Keypoint Detection}: ORB utilizes FAST (Features from Accelerated Segment Test) for efficient keypoint detection.
%     \item \textit{Keypoint Descriptor}: Descriptors are created using BRIEF (Binary Robust Independent Elementary Features), which is lightweight and computationally efficient.
% \end{itemize}
\textbf{IntFeat (Proposed Method)}: The \textbf{IntFeat} feature detection method presents a robust approach by integrating the strengths of SIFT's scale invariance and ORB's computational efficiency.
\begin{algorithm}[H]
\caption{IntFeat Algorithm for Image Registration}
\begin{algorithmic}
\Procedure{Intfeat}{$img1$, $img2$}

    \State \textbf{EXTRACT} SIFT and ORB key points and descriptors from both $img1$ and $img2$
    \State \textbf{COMBINE} SIFT and ORB key points for both images
    \State \textbf{REDUCE} the dimensionality of SIFT descriptors to match ORB using Principal Component Analysis (PCA)
    \State Use \textbf{BRUTE-FORCE} matcher to find the nearest matches between the descriptors of both images
    \State \textbf{COMPUTE} homography using the RANSAC algorithm based on matched key points
    \State \textbf{APPLY} perspective transformation using the homography matrix to align the images
    \State \Return The registered image
\EndProcedure
\end{algorithmic}
\end{algorithm}

\subsubsection{Feature Matching}
\textbf{BF (Brute-Force) Matcher}
In the BF matcher, each feature descriptor from the first set is compared to all feature descriptors in the second set using a chosen distance metric, and the closest match is selected.
\subsubsection{Transformation Model}
The homography matrix represents the transformation between two planes. It is a 3x3 matrix with 8 degrees of freedom, estimated using the RANSAC (RANdom SAmple Consensus) algorithm to ensure robustness against outliers.

\subsection{SyncVision}
SyncVision is a Python package for image processing that allows users to compare two images using different image registration methods. The package supports SIFT, ORB, and IntFeat registration methods and provides metrics such as PSNR and SSIM to evaluate the quality of the registered images.
\section{Results}
%%%%%%%%%%%%%%%%%%%%%%%%%%%%%%%%%%%%%%%%%%%%%%%%%%%%%%%%%%%%%%%%%%%%%%%%%%%%%%%%%%%%%%%%%%%
\subsection{Performance of Feature Detectors}
Table\ref{tab1:results_before} compares the performance of different feature detection methods (SIFT, ORB, and IntFeat) using two interpolation techniques (bi-linear and bi-cubic) based on SSIM (Structural Similarity Index Measure) and PSNR (Peak Signal-to-Noise Ratio) metrics. IntFeat shows a balanced performance between SIFT and ORB. For bi-linear interpolation, IntFeat achieves a slightly lower SSIM (0.7633) than SIFT but outperforms ORB. For bi-cubic interpolation, IntFeat’s performance drops further with a lower SSIM (0.7380) and PSNR (27.781), indicating that bi-linear interpolation is more effective for IntFeat than bi-cubic in this case.

\begin{table}[htbp]
\caption{Comparison of Feature Extraction Methods for Bi-linear and Bi-cubic interpolation on lunar images}
\begin{center}
\begin{tabular}{|c|c|c|c|c|}
\hline
\textbf{Interpolation} & \multicolumn{2}{|c|}{\textbf{Bi-linear}} & \multicolumn{2}{|c|}{\textbf{Bi-cubic}} \\
\hline
\textbf{Evaluation Metric} & \textbf{\textit{SSIM}} & \textbf{\textit{PSNR}} & \textbf{\textit{SSIM}} & \textbf{\textit{PSNR}} \\
\hline
SIFT & 0.7694 & 29.134 & 0.7585 & 29.134 \\
\hline
ORB & 0.7554 & 28.404 & 0.7475 & 28.268 \\
\hline
\textbf{IntFeat} & \textbf{0.7633} & \textbf{28.774} & \textbf{0.7380} & \textbf{27.781} \\
\hline
\end{tabular}
\label{tab1:results_before}
\end{center}
\end{table}
%%%%%%%%%%%%%%%%%%%%%%%%%%%%%%%%%%%%%%%%%%%%%%%%%%%%%%%%%%%%%%%%%%%%
Table\ref{tab2:results_after} presents a similar comparison but evaluated on high-angle lunar images. SIFT demonstrates the highest performance in both interpolation methods, achieving SSIM values of 0.7887 and 0.7825, along with corresponding PSNR values of 29.645 and 30.005. ORB follows, with lower SSIM and PSNR values. IntFeat closely matches SIFT’s performance in bi-linear interpolation (SSIM: 0.7880, PSNR: 29.600) and remains competitive in bi-cubic interpolation (SSIM: 0.7812, PSNR: 29.978), indicating its effectiveness for high-angle lunar images.
\begin{table}[htbp]
\caption{Comparison of Feature Extraction Methods for Bi-linear and Bi-cubic interpolation on high-angle lunar images}
\begin{center}
\begin{tabular}{|c|c|c|c|c|}
\hline
\textbf{Interpolation} & \multicolumn{2}{|c|}{\textbf{Bi-linear}} & \multicolumn{2}{|c|}{\textbf{Bi-cubic}} \\
\hline
\textbf{Evaluation Metric} & \textbf{\textit{SSIM}} & \textbf{\textit{PSNR}} & \textbf{\textit{SSIM}} & \textbf{\textit{PSNR}} \\
\hline
SIFT & 0.7887 & 29.645 & 0.7825 & 30.005 \\
\hline
ORB & 0.7750 & 29.200 & 0.7653 & 28.945 \\
\hline
\textbf{IntFeat} & \textbf{0.7880} & \textbf{29.600} & \textbf{0.7812} & \textbf{29.978} \\
\hline
\end{tabular}
\label{tab2:results_after}
\end{center}
\end{table}
\begin{figure}
    \centering
    \includegraphics[width=\linewidth]{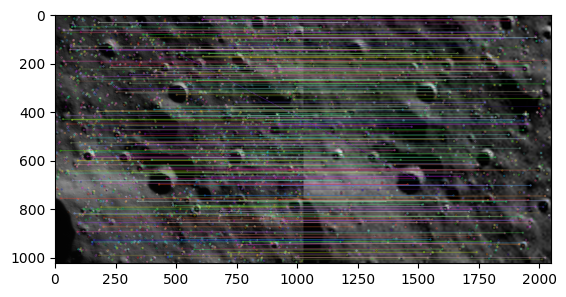}
    \caption{Image registration result between a TMC-2 and OHRC patch, demonstrating feature matching via keypoints and descriptors. The x and y axes indicate pixel coordinates on a 1024x1024 scale.}
    \label{fig:image_registration}
\end{figure}
%%%%%%%%%%%%%%%%%%%%%%%%%%%%%%%%%%%%%%%%%%%%%%%%%%%%%%%%%%%%%%%%%%%%%
\subsection{Analysis of Lunar Images}
During the initial analysis, it was observed that lunar images taken from low angles relative to the sun exhibited high contrast and sharp shadows. This made the images challenging to upscale using standard interpolation methods, such as bi-linear and bi-cubic interpolation, as both methods resulted in reduced quality, particularly in terms of SSIM and PSNR. The upscaling process worsened image quality due to the exaggerated contrast and shadow artifacts as shown in Table \ref{tab1:results_before}.\\[5pt]
%%%%%%%%%%%%%%%%%%%%%%%%%%%%%%%%%%%%%%%%%%%%%%%%%%%%%%%%%%%%%%%%%%%%%%%%
%%%%%%%%%%%%%%%%%%%%%%%%%%%%%%%%%%%%%%%%%%%%%%%%%%%%%%%%%%%%%%%%%%%%%
To address this, we examined the satellite metadata and filtered the images based on the angle of the sun during image capture. By selecting only those images taken at higher angles relative to the sun, we were able to mitigate the issues caused by extreme contrast and shadowing. This selection led to a marked improvement in the performance of both interpolation methods as shown in Table \ref{tab2:results_after}.\\[5pt]
%%%%%%%%%%%%%%%%%%%%%%%%%%%%%%%%%%%%%%%%%%%%%%%%%%%%%%%%%%%%%%%%%%
Interestingly, while bi-cubic interpolation initially performed worse on low-angle images, it demonstrated superior performance on high-angle images, as expected, due to its ability to preserve finer details during upscaling. This improvement highlights the importance of considering the angle of illumination in lunar image processing for achieving optimal interpolation results.
%%%%%%%%%%%%%%%%%%%%%%%%%%%%%%%%%%%%%%%%%%%%%%%%%%%%%%%%%%%%%%%%%%%%%%%%%%%%%
\section{Discussions}
Adapting to the unique characteristics of lunar imagery presents significant challenges, primarily due to sharp shadow boundaries and high-contrast regions that lead to substantial brightness variations. These conditions complicate the effectiveness of feature detection algorithms. One concern is the noise levels introduced by the IntFeat detector. The low-level feature extraction of ORB tends to amplify noise, particularly in the form of artifacts. This amplification is exacerbated during the smoothing of textural details in the image interpolation process, impacting the overall performance.\\[5pt]
Additionally, IntFeat’s limitations are highlighted by its struggle to handle the combination of high noise levels, extreme contrast differences, homogenous textures, and fluctuating lighting conditions. These factors collectively hinder the detector’s ability to effectively integrate both low- and high-level features, resulting in sub-optimal performance in challenging lunar environments.
\section{Conclusions \& Future Work}
In general image processing tasks, bicubic interpolation typically outperforms bilinear interpolation due to its reliance on a larger 4x4 neighborhood for smoother transitions. However, this advantage does not carry over to lunar images, where the presence of inherent noise and sharp contrast boundaries limits its effectiveness. The custom feature detector, IntFeat, demonstrated performance comparable to the baseline methods but did not exceed the accuracy of existing detectors.\\[5pt]
The current implementation highlights the need for further refinement, particularly in filtering key points and developing more robust descriptors that can better handle the unique challenges of lunar imagery, such as high-contrast regions and elevated noise levels. Future work will focus on improving these aspects to enhance feature detection in the demanding environment of lunar surfaces.
\newpage

% \section{References}

\appendix
\subsection{Structural Similarity Index Measure (SSIM)}
\label{ssim}
The performance of dehazing methods is evaluated by the following equation of SSIM score between two images:
\begin{equation}
    SSIM(x, y) = \frac{(2 \mu_x \mu_y + c_1)(2 \sigma_{xy} + c_2)}{(\mu_x^2 + \mu_y^2 + c_1)(\sigma_x^2 + \sigma_y^2 + c_2)}
\end{equation}

where:
\begin{itemize}
    \item $\mu_x$ and $\mu_y$ are the average of $x$ and $y$ respectively.
    \item $\sigma_x^2$ and $\sigma_y^2$ are the variance of $x$ and $y$ respectively.
    \item $\sigma_{xy}$ is the covariance of $x$ and $y$.
    \item $c_1 = (k_1 L)^2$ and $c_2 = (k_2 L)^2$ are two variables to stabilize the division with weak denominator; $L$ is the dynamic range of the pixel-values (typically this is $2^{bits\_per\_pixel} - 1$), $k_1=0.01$ and $k_2=0.03$ by default.
\end{itemize}
\subsection{Peak Signal-to-Noise ratio (PSNR)}
\label{psnr}
Peak Signal-to-Noise Ratio (PSNR) is a widely used metric for evaluating the quality of reconstructed images or videos compared to the original, reference data. It is expressed in decibels (dB) and is calculated based on the mean squared error (MSE) between the original and the reconstructed images. The formula for PSNR is given by:

\begin{equation}
\text{PSNR} = 10 \cdot \log_{10} \left( \frac{\text{MAX}^2}{\text{MSE}} \right),
\end{equation}

where \(\text{MAX}\) is the maximum possible pixel value of the image (for example, 255 for 8-bit images), and \(\text{MSE}\) is the mean squared error, defined as:

\begin{equation}
\text{MSE} = \frac{1}{mn} \sum_{i=0}^{m-1} \sum_{j=0}^{n-1} \left( I(i,j) - K(i,j) \right)^2,
\end{equation}

where \(I(i,j)\) represents the pixel value at position \((i,j)\) in the original image, and \(K(i,j)\) represents the pixel value at the same position in the reconstructed image. Higher PSNR values generally indicate better reconstruction quality, as they imply a lower MSE and thus less distortion. PSNR is particularly useful for comparing the performance of different image processing algorithms in tasks such as image compression, denoising, and super-resolution.
\subsubsection{Mean Average Precision (mAP)}
For object detection performance, we are using mean Average Precision (mAP):
\begin{equation}
    AP = \frac{\sum_{k=1}^{n} (P(k) \times \text{rel}(k))}{\text{number of relevant documents}}
\end{equation}

where:
\begin{itemize}
    \item $P(k)$ is the precision at cutoff $k$ in the list.
    \item $\text{rel}(k)$ is an indicator function equaling 1 if the item at rank $k$ is a relevant document, 0 otherwise.
    \item $n$ is the number of retrieved documents.
\end{itemize}

The mean Average Precision is then calculated as:

\begin{equation}
    mAP = \frac{\sum_{q=1}^{Q} AP_q}{Q}
\end{equation}

where $AP_q$ is the Average Precision for the $q^{th}$ query and $Q$ is the total number of queries.
\end{document}